\documentclass[runningheads]{llncs}
\usepackage[T1]{fontenc}
\usepackage{graphicx}
\usepackage{multirow}
\usepackage{algorithm}
\usepackage{algorithmic}
\usepackage{amsmath}
\usepackage{amssymb}
\usepackage{subcaption}
\usepackage{dsfont}
\usepackage{mathrsfs}
\usepackage[dvipsnames]{xcolor}
\usepackage{tikz}

\definecolor{lightblue}{RGB}{173, 216, 230}
\begin{document}
\title{A Generalized Apprenticeship Learning Framework for Capturing Evolving Student Pedagogical Strategies}
\titlerunning{AL Framework for Capturing Evolving Student Pedagogical Strategies}
% \author{}
\author{Md Mirajul Islam\inst{1}\orcidID{0000-0001-5086-7399} \and
Xi Yang\inst{2}\orcidID{0000-0003-0026-9096} \and
Rajesh Debnath\inst{1}\orcidID{0000-0002-5460-9773} \and
Adittya Shoukarjya Saha\inst{1}\orcidID{0000-0001-6344-9663} \and
Min Chi\inst{1}\orcidID{0000-0003-1765-7837}}
\authorrunning{M. M. Islam et al.}
\institute{North Carolina State University, Raleigh, NC 27606, USA\\
\email{\{mislam22, rdebnat, asaha4, mchi\}@ncsu.edu}
\and
IBM Research, Yorktown Heights, NY 10598, USA\\
\email{xi.yang@ibm.com}}
\maketitle              % typeset the header of the contribution
\begin{abstract}
% old abstract - submitted paper
% Apprenticeship Learning leverages \emph{expert demonstrations} to infer the \textbf{expert's underlying reward functions} and derive decision-making policies that generalize and replicate optimal behavior. In this work, we leveraged \emph{expert student demonstrations} to extract effective \emph{pedagogical decision-making policies}, which are essential for maximizing  the full potential of adaptive educational technologies like Intelligent Tutoring Systems. More specifically, we applied a generalized Apprenticeship Learning framework to induce effective pedagogical policies by capturing the complexities of the expert student learning process, where \textbf{multiple reward functions may dynamically evolve over time}. We evaluate the effectiveness of THEMES against six state-of-the-art baselines, demonstrating its superior performance and highlighting its potential as a powerful alternative for inducing effective pedagogical policies. 
%---------------------------------------------
% new abstract - camera ready
Reinforcement Learning (RL) and Deep Reinforcement Learning (DRL) have advanced rapidly in recent years and have been successfully applied to e-learning environments like intelligent tutoring systems (ITSs). Despite great success, the broader application of DRL to educational technologies has been limited due to major challenges such as \emph{sample inefficiency} and difficulty designing the \emph{reward function}. In contrast, Apprenticeship Learning (AL) uses a few \emph{expert demonstrations} to infer the \textbf{expert's underlying reward functions} and derive decision-making policies that generalize and replicate optimal behavior. In this work, we leverage a generalized AL framework, THEMES, to induce effective pedagogical policies by capturing the complexities of the expert student learning process, where \textbf{multiple reward functions may dynamically evolve over time}. We evaluate the effectiveness of THEMES against six state-of-the-art baselines, demonstrating its superior performance and highlighting its potential as a powerful alternative for inducing effective pedagogical policies and show that it can achieve high performance, with an AUC of 0.899 and a Jaccard of 0.653, using only 18 trajectories of a previous semester to predict student pedagogical decisions in a later semester.


\end{abstract}
%
%
%
%%%%%%%%%%%%%%%%%%%%%%%%%%%%%%%%%%%%%%%%%%%%%%%%%%%%%%%%%%%%%%%%%%%%%%%%%%%%%%%%%%%%%%%%%%%%%%%%%%%%
\section{Introduction}
\vspace{-0.1in}
% Dr. Chi's suggestions on writing Introduction
% \textit{***Follow IJCAI25's paper:
% 1. AL  ....  Compared with RL,  a short summary of your old EDM paper : data efficient (references)  and reward function (references).
% 2. Previous work on AL in the ITSs: .... *Mirajul's literature review in EDM24 paper + your own EDM24)
% 3. Despite prior success, prior work of AL in education/iTSs faces two major challenges: 1) XXXX assumes all demonstrations are generated from a homogeneous policy driven by a single reward function. In practice, however, clinicians adaptively implement different policies with multiple reward functions evolving over time according to the patient's disease progression stage (\textit{e.g.}, common fever vs. severe shock) \cite{wang2020multi,wang2022hierarchical}. To address this, we propose a hierarchical AL that automatically partitions trajectories (\textit{i.e.}, state-action pairs) into sub-trajectories, accounting for the varying reward functions. 2) EM-IRL and EM-EDM  assumes all demonstrations are collected at regular time intervals. However, real-world data, such as electronic health records, are generally collected irregularly, with time intervals ranging from seconds to days \cite{baytas2017patient}. To address this, we develop a time-aware AL framework capable of capturing latent progressive patterns to induce more accurate policies. identify student's pedagogical action***}

%%%%%%%%%%%%%%%%%%%%%%%%%%%%%%%%%%%%%%%%%%%%%%%%%%%%%%%%%%%%%%%%%%%%%%%%%%%%%%%%%%%%%%%%%%%%%%%%%%%%
%% Part 1
In most STEM e-learning environments, the system follows a sequential decision-making process, deploying a \emph{pedagogical policy} to determine \emph{what} action to take and \emph{when} at each discrete time step. In such environments, student-system interactions can be modeled as sequential decision-making problems under uncertainty, formulated within the framework of Reinforcement Learning (RL)—a paradigm that optimizes long-term rewards without requiring knowledge of the ‘best’ decisions at each immediate time step \cite{sutton2018reinforcement}. Recent research has demonstrated the effectiveness of RL and Deep RL (DRL) in deriving data-driven pedagogical policies to enhance student learning in Intelligent Tutoring Systems (ITSs) (e.g., \cite{abdelshiheed2023bridging,ju2020aiedcritical,mandel2014offline,rowe2015improving}). Despite their success, several challenges hinder its broader application in e-learning systems. One major challenge is \textbf{sample inefficiency}, as classic DRL algorithms like Deep Q-Networks (DQN) require millions of interactions to learn effective policies. Many studies attempt to derive pedagogical policies from fewer than 3,000 student-ITS interactive  trajectories, often leading to ineffective results \cite{DBLP:conf/aied/AusinMBC20,shen2016reinforcement}. Another key challenge is \textbf{reward function} design, which is crucial for guiding RL agents but is difficult to define accurately, especially in human-centric domains like education \cite{kaelbling1996reinforcement}. Manually designing reward functions is labor-intensive, prone to expertise blind spots, and often results in misspecified objectives that misalign with intended policies \cite{goyal2019using,abbeel2004apprenticeship,amodei2016concrete}.

% \vspace{-0.1in}
Unlike RL and DRL, \textbf{Apprenticeship Learning (AL)} is a machine learning approach that derives decision-making policies from a limited set of \emph{expert demonstrations} by inferring the \textbf{expert's underlying reward functions}, enabling the system to generalize and replicate optimal behavior \cite{abbeel2004apprenticeship}. It assumes expert agents make \emph{optimal} or \emph{near-optimal} decisions based on an underlying reward function, allowing the apprentice to efficiently approximate expert behavior with fewer training samples \cite{abbeel2004apprenticeship,ziebart2008maximum}.  Existing AL approaches are typically online, requiring iterative interactions with the environment to collect new data and update the model \cite{abbeel2004apprenticeship,ziebart2008maximum,ho2016generative,finn2016guided}. However, executing a potentially flawed policy can be costly or even dangerous in human-centric tasks such as education \cite{levine2020offline}, making offline policy induction from expert demonstrated highly desirable \cite{asoh2013application,pan2019dissecting,rafferty2015inferring,wang2020inferring,islam2024generalized}. In this work,  \textbf{our "expert" demonstrations} were selected by allowing students to make pedagogical decisions, and then choosing only those in which the student greatly benefited from the learning environment.  To accurately model students' complex and dynamic decision-making processes, it is crucial to incorporate \emph{evolving reward functions} that reflect students' changing learning needs and adaptive strategies across different learning phases. Previous offline AL methods typically assume a single or fixed reward function across trajectories, failing to capture the evolving goals of students. For example, while energy-based distribution matching (EDM) \cite{jarrett2020strictly} has advanced offline AL, it assumes a homogeneous policy with a single reward function \cite{babes2011apprenticeship}. Approaches like EM-IRL \cite{yang2020student} address reward function heterogeneity by clustering students based on pedagogical behavior but struggle with large continuous state spaces. More recently, EM-EDM \cite{islam2024generalized} extends EDM with an Expectation-Maximization(EM) framework, enabling policy induction from expert demonstrations driven by multiple heterogeneous reward functions while generalizing to continuous state spaces. 

To directly induce effective pedagogical policies from expert students' interactions with an ITS,  we leverage \textbf{THEMES} \cite{yang2023dpm}, a \textbf{T}ime-aware \textbf{H}ierarchical \textbf{EM} \textbf{E}nergy-based \textbf{S}ubtrajectory AL framework which is designed to model \emph{multiple} reward functions \emph{evolving} over time in an \emph{offline} manner. THEMES  consists of two key components: \textbf{(1) Sub-trajectory partitioning}, achieved by partitioning expert trajectories into fine-grained sub-trajectories while incorporating time-awareness. A hierarchically learned reward regulator captures evolving decision-making patterns. \textbf{(2) Policy induction}, performed through offline EM-EDM, which simultaneously clusters sub-trajectories and induces policies specific to each cluster. We evaluate THEMES against six competitive AL baselines and two ablations on a challenging task—predicting students' pedagogical decisions in an ITS by training on past semester trajectories and testing on a later semester. Results show that THEMES consistently outperforms all baselines and ablations across performance metrics. Our findings highlight its potential as a powerful alternative for capturing evolving student pedagogical strategies and inducing effective policies, even with as few as 18 demonstrations.
\vspace{-0.1in}
\section{Related Work}
\vspace{-0.1in}
\subsubsection{RL for Pedagogical Policy Induction:}
Prior research has demonstrated the great potential of leveraging RL and DRL for pedagogical policy induction in ITSs \cite{ju2020aiedcritical,mandel2014offline,rowe2015improving,wang2017interactive}. Shen et al. \cite{shen2016reinforcement} applied a value iteration algorithm to induce policies aimed at improving student performance, showing advantages over random decision-making. Wang et al. \cite{wang2017interactive} utilized DRL to optimize students' learning in an educational game, outperforming traditional RL approaches in simulations. More recently, Zhou et al. \cite{zhou2019hierarchical} employed offline Hierarchical RL to induce pedagogical policies, achieving superior results compared to flat-RL baselines in classroom studies. However, these approaches require large-scale training data—often exceeding 1,000 trajectories \cite{shen2016reinforcement,ju2020aiedcritical,zhou2019hierarchical}—which is typically collected by constraining ITSs to make randomized yet reasonable decisions. Moreover, they rely on predefined reward functions, such as normalized learning gains \cite{shen2016reinforcement,zhou2019hierarchical}, manually designed by researchers \cite{goyal2019using}, limiting adaptability to diverse student learning processes.
\vspace{-0.2in}
\subsubsection{Offline Apprenticeship Learning (AL):} The most straightforward approach to \emph{offline} AL is behavior cloning (BC) \cite{gleave2022imitation}, which directly maps states to actions to replicate demonstrated behaviors \cite{pomerleau1991efficient}. More robust alternatives, such as inverse reinforcement learning (IRL) \cite{klein2011batch,jain2019model,lee2019truly,chan2021scalable} and adversarial imitation learning \cite{kostrikov2018discriminator,kostrikov2019imitation}, infer an explicit or implicit reward function to generalize beyond observed behaviors. Standard IRL involves iterative online computations, including reward function inference, policy induction via RL, policy rollouts, and reward updates based on behavior divergence. To eliminate policy rollouts, batch-IRL methods \cite{raghu2017continuous} have been introduced, but they often rely on off-policy evaluation, leading to suboptimal solutions. Similarly, adversarial imitation learning iteratively trains a generator to roll out policies and a discriminator to distinguish learned behaviors from expert demonstrations. Off-policy variants based on actor-critic algorithms \cite{kostrikov2018discriminator,kostrikov2019imitation} mitigate the need for policy rollouts but inherit the instability of adversarial training \cite{ho2016generative}. To address these challenges, EDM \cite{jarrett2020strictly} was introduced, offering a more stable solution for offline settings and outperforming both IRL and adversarial imitation learning. In this work, \emph{EDM serves as one of our baselines}. All the aforementioned methods assume a \textit{single} reward function across all demonstrations.

To handle scenarios where \textit{multiple} reward functions exist across trajectories but \emph{remain fixed within each trajectory}, several methods have been developed. Dimitrakakis and Rothkopf~\cite{dimitrakakis2011bayesian} introduced Bayesian multi-task IRL, which learns a separate reward function for each trajectory while sharing a common prior distribution. Choi et al.\cite{choi2012nonparametric} extended this idea using nonparametric Bayesian IRL, modeling the distribution of different rewards with a Dirichlet process to allow for more flexible clustering. Babes-Vroman et al.\cite{babes2011apprenticeship} proposed an EM-based framework that iteratively assigns probabilities to demonstrations belonging to different clusters and updates the corresponding cluster-wise reward functions using Maximum Likelihood IRL.

To model reward functions that \emph{evolve} over time, Krishnan et al. \cite{krishnan2016hirl} introduced Hierarchical IRL (HIRL), and Hausman et al. \cite{hausman2017multi} proposed Multi-modal Imitation Learning. More recently, Wang et al.~\cite{wang2020multi,wang2022hierarchical} developed a hierarchical imitation learning model that learns both high-level policies and sub-policies for individual sub-tasks. These methods, often relying on generative adversarial imitation learning \cite{ho2016generative}, require online interactions or off-policy evaluations, making them less suited for fully offline settings. Additionally, meta-imitation learning \cite{finn2017one} has been explored to generalize from a few demonstrations, typically by structuring tasks into subtasks predefined by domain experts. However, manually defining subtasks is challenging in complex, uncertain environments like education. To address this, THEMES framework \cite{yang2023dpm} \emph{automatically segments trajectories into sub-trajectories}, capturing the evolving nature of reward functions in a data-driven manner.

% To model reward functions that \emph{evolve} over time, Krishnan et al.\cite{krishnan2016hirl} introduced Hierarchical IRL (HIRL), which decomposes long-horizon tasks into sub-tasks based on consistent transitions observed in demonstrations. Similarly, Hausman et al.\cite{hausman2017multi} proposed Multi-modal Imitation Learning, which jointly segments and imitates policies by modeling time-dependent policy heterogeneity using a latent intention variable within a generative adversarial imitation learning framework. More recently, Wang et al.~\cite{wang2020multi,wang2022hierarchical} developed a hierarchical imitation learning model that learns both high-level policies and sub-policies for individual sub-tasks. These methods, often relying on generative adversarial imitation learning \cite{ho2016generative}, require online interactions or off-policy evaluations, making them less suited for fully offline settings. Additionally, meta-imitation learning \cite{finn2017one} has been explored to generalize from a few demonstrations, typically by structuring tasks into subtasks predefined by domain experts. However, manually defining subtasks is challenging in complex, uncertain environments like education. To address this, THEMES framework \emph{automatically segments trajectories into sub-trajectories}, capturing the evolving nature of reward functions in a data-driven manner.
 
\vspace{-0.2in}
\subsubsection{AL for ITSs:}
Rafferty et al. applied IRL to infer learners' beliefs in an educational game, demonstrating that IRL could effectively recover participants' understanding of how their actions influence the environment \cite{rafferty2015inferring}. This highlighted IRL’s potential for interpreting data in interactive learning settings. In subsequent work, they used IRL to assess learners’ mastery of algebraic skills, identifying misconceptions and providing personalized feedback for improvement \cite{rafferty2016using}. Yang et al. introduced EM-Inverse Reinforcement Learning (EM-IRL) to cluster students based on pedagogical behavior \cite{yang2020student}. However, like many multi-intent approaches, it was constrained to discrete state representations (e.g., 17 discrete features in \cite{yang2020student}). More recently, Islam et al. developed EM-EDM framework \cite{islam2024generalized} to induce effective policies from expert demonstrations with heterogeneous reward functions while generalizing to continuous state spaces. They demonstrated that EM-EDM surpassed two DRL and four AL baselines in a challenging student modeling task for predicting pedagogical decisions. Thus, we adopt EM-EDM as a baseline to assess the effectiveness of THEMES.

% A few AL methods have been proposed to consider multiple evolving reward functions. A Bayesian multi-task IRL was proposed, which models the heterogeneity of reward functions by formalizing it as a statistical preference elicitation via a joint reward-policy prior \cite{dimitrakakis2011bayesian}. Choi and Kim integrated a Dirichlet process mixture model into Bayesian IRL to cluster the demonstrations \cite{choi2012nonparametric}. Using a Bayesian model, they incorporated the domain knowledge of multiple reward functions. Similarly, Arora \textit{et al.} combined the Dirichlet process with a maximum entropy IRL to learn the clusters of demonstration with different reward functions \cite{arora2021min}.

% Babes \textit{et al.} derived an EM-based IRL approach that clusters trajectories based on their different reward functions \cite{babes2011apprenticeship}. As a component of the EM-based IRL, a maximum-likelihood IRL uses a gradient ascent method to optimize the reward parameters, which has successfully identified unknown reward functions. Xi et al. used EM-Inverse Reinforcement Learning (EM-IRL) to subtype students from their pedagogical behavior data \cite{yang2020student}. Still, as most other multiple intentions works, it is limited because it can only handle discrete state representation (17 discrete features involved in \cite{yang2020student}) and can not generalize to large continuous state space.

%%%%%%%%%%%%%%%%%%%%%%%%%%%%%%%%%%%%%%%%%%%%%%%%%%%%%%%%%%%%%%%%%%%%%%%%%%%%%%%%%%%%%%%%%%%%%%%%%%%%
\vspace{-0.15in}
\section{Methods}
\begin{figure}
\vspace{-0.2in}
    \centering
    \includegraphics[scale=0.4]{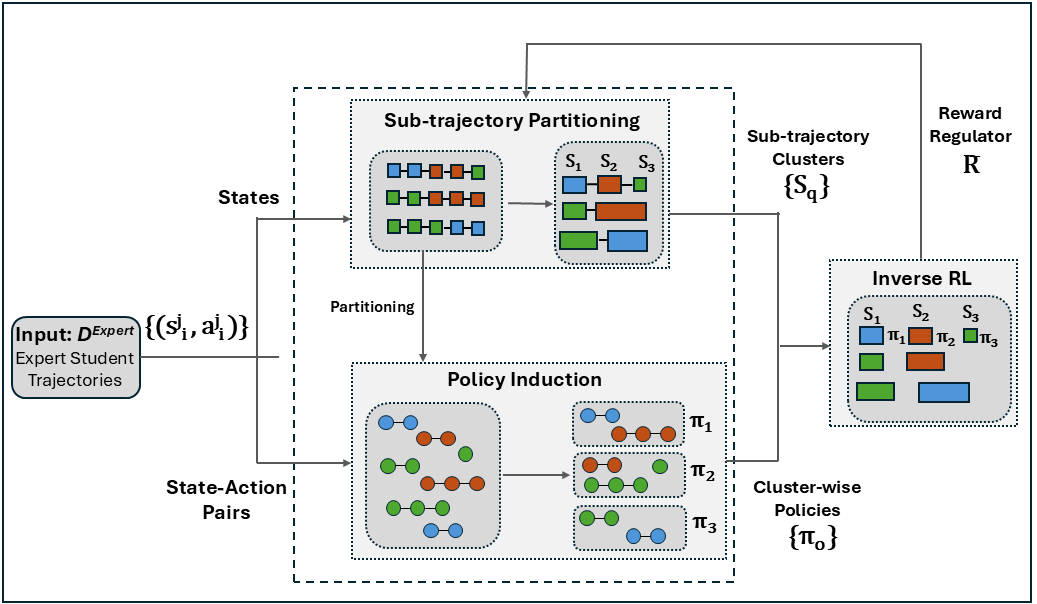}
    \caption{Overview of the THEMES Framework.}
\label{fig:overview}
\vspace{-0.1in}
\end{figure}
\vspace{-0.10in}

% \vspace{-0.2in}
% \subsubsection{Preliminaries}

The input dataset $\mathcal{D}^{Total}$ consists of $L$ student-ITS interaction trajectories, each having a fixed length of $n$. Each trajectory $d$ can be viewed as: $s_{1} \xrightarrow{a_{1}} s_{2} \xrightarrow{a_{2}} \cdots s_{n-1} \xrightarrow{a_{n-1}} s_{n}$. Here $s_{i} \xrightarrow{a_{i}} s_{i+1}$ designates that at the $i_{th}$ time step in $d$, the student's learning status was in state $s_{i}$, the student executed the action $a_{i}$ and their learning status transitioned to state $s_{i+1}$. Since in AL, it is typically assumed that the experts' trajectories provided as input are optimal or near-optimal according to latent reward functions \cite{abbeel2004apprenticeship}, the quality of these trajectories is crucial for inducing more accurate policies. Therefore, we have designed a procedure outlined in Section \ref{TrajSelection} to select the expert student trajectories, based on which a subset of $N$ trajectories $\mathcal{D}^{Expert}$ is selected from $\mathcal{D}^{Total}$, denoted as: $\{\mathcal{D}^{j}\} = \{(\mathbf{s}_i^j, a_{i}^{j})|i=1, ..., n; j = 1,..., N\}$ where $N \le L$.

\subsection{THEMES for Evolving Student Pedagogical Strategy}
\label{subsec:themes}
% The overview of THEMES is illustrated in Figure \ref{fig:overview}, with additional details provided in Algorithm \ref{alg:themes}. The input consists of $N$ trajectories demonstrated by experts, each composed of $\{(\mathbf{x}_t^n, a_{t}^{n})|t \in [1,T^{n}]\}$, $n \in [1,N]\}$, where $\mathbf{x}_t^n$ represents the state at the $t$-th timestamp, $a_{t}^{n}$ is the corresponding action, and $T^{n}$ denotes the length of the $n$-th trajectory.

% \vspace{-0.15in}
% \subsubsection{Overview}
Given $\mathcal{D}^{Expert}$ as input, THEMES seeks to learn the multiple underlying reward functions of students that evolve over time leveraging its two key components, \textit{i.e.}, sub-trajectory partitioning and policy induction (Figure \ref{fig:overview}).
Primarily, a time-aware sub-trajectory partitioning method \cite{yang2021multi} processes the states from input ($\mathcal{D}^{Expert}$). It partitions and clusters the states $\{\mathbf{s}_i^j\}$ into sub-trajectories such that each resulting cluster captures consistent time-invariant patterns, which can be interpreted as high-level states $\{S_{q} | q \in [1,Q]\}$ where Q refers to the number of sub-trajectory clusters. Then, focusing on the state-action pairs within the partitioned sub-trajectories, an Expectation-Maximization(EM) Energy-based Distribution Matching (EDM) method \cite{jarrett2020strictly} clusters and induces policies from the state-action pairs $\{(\mathbf{s}_i^j, a_{i}^{j})\}$ over the partitioned sub-trajectories, ensuring that each cluster exhibits consistent decision-making patterns, referred to as high-level actions $\{\pi_{o} | o \in [1,O]\}$. Using these high-level state-action pairs, a Maximum Likelihood Inverse Reinforcement Learning (ML-IRL) method \cite{babes2011apprenticeship} learns a high-level reward regulator $\overline{R}(\cdot)$, which is then fed back to refine sub-trajectory partitioning. This iterative procedure continues until convergence.

\vspace{-0.1in}
\subsubsection{Sub-trajectory Partitioning}
\label{sec:rmt-ticc}
% \subsubsection{Preliminaries} 
\vspace{-0.1in}
% The \textit{high-level states} are extracted by simultaneously partitioning and clustering trajectories based on their latent patterns. More specifically, each state $\mathbf{s}_i^j$ is assigned to a cluster $\{p|p \in [1,P]\}$. Since states in a trajectory are consecutive and dependent on their neighbors, rather than treating each state independently, THEMES take advantage of temporal dependencies by exploring patterns within a sliding window of size $\omega \ll n$. For a given state $\mathbf{s}_i^j$, the context within the window, denoted as $\mathbf{S}_i^j=\{\mathbf{s}_{i-\omega+1}^{j},...,\mathbf{s}_i^j\}$, is considered to determine which cluster $\mathbf{s}_i^j$ belongs to. To learn unsupervised cluster mappings, each sliding window $\mathbf{S}_i^j$ is treated as a $m\omega$-dimensional random variable (concatenating the $\omega$ timestamps in $\mathbf{S}_i^j$). All states (characterized by sliding windows) are then optimally fitted into $P$ Gaussian distributions, with the $p$-th distribution corresponding to the $p$-th sub-trajectory cluster.

As shown in Figure \ref{fig:overview}, this module first extracts the states $\{\mathbf{s}_i^j\}$ from the given expert student trajectories $\{(\mathbf{s}_i^j, a_i^j)\}$. 
% the \textit{high-level states} are identified by jointly partitioning and clustering trajectories based on their underlying latent patterns. Specifically, each state $\mathbf{s}_i^j$ is assigned to a sub-trajectory cluster $\{q \mid q \in [1, Q]\}$.
Since the states in student trajectories are sequential and influenced by adjacent states, it is desirable to leverage the temporal dependencies rather than treating states independently. Hence, the \emph{Subtrajectory-partitioning} module of THEMES captures this temporal dependency by investigating patterns within a sliding window of size $\omega \ll n$. For each state $\mathbf{s}_i^j$, the sub-trajectory within the window, $\mathbf{S}_i^j = \{\mathbf{s}_{i-\omega+1}^{j}, \ldots, \mathbf{s}_i^j\}$, is used to determine the cluster assignments $\{S_{q} | q \in [1,Q]\}$. Given $\mathbf{s}_i^j \in \mathbb{R}^m$, each sub-trajectory with the sliding window $\mathbf{S}_i^j$ can be viewed as a random variable in a $m\omega$-dimensional space, which can be generated by concatenating the $\omega$ consecutive states. 
% Finally, all states {$\mathbf{s}_i^j$} are optimally fitted to $Q$ sub-trajectory clusters, where the $q$-th cluster represents $q$-th Gaussian distribution.

Each sub-trajectory cluster is represented by its mean vector ($\mathbf{\mu}$) and inverse covariance matrix ($\mathbf{\Theta}$) \cite{hallac2017toeplitz}. Specifically, the process of determining the optimal mean vectors, $\{\mathbf{\mu}_{q} \mid q \in [1,Q]\}$, is equivalent to assigning each state to the most suitable cluster. In Intelligent Tutoring Systems (ITSs), student trajectories are often collected at irregular intervals. Consequently, both the mean vectors and inverse covariance matrices are learned across various trajectories, while \textit{time-awareness} is incorporated during the sub-trajectory partitioning phase using a decay function that allows for a re-evaluation of cluster consistency over time \cite{yang2021multi}. To enhance the sub-trajectory clustering by accounting for decision-making behaviors, \emph{hierarchical} rewards $\{\overline{r}_{t}^{j} \mid t \in [1,n]\}$ are inferred for each state-action pair, which then inform the adjustment of the consistency constraint. As illustrated in Figure \ref{fig:overview}, these high-level state-action pairs facilitate the inference of high-level rewards, $\overline{R}(\cdot)$, which can be learned using Maximum Likelihood Inverse Reinforcement Learning (ML-IRL) \cite{babes2011apprenticeship}.

\vspace{-0.2in}
\subsubsection{Policy Induction}
In this segment, EDM \cite{jarrett2020strictly}, a state-of-the-art strictly offline AL method, serves as the core component, which assumes a single reward function while learning the policy $\pi^\theta$, parameterized by $\theta$. To induce the policy $\pi^\theta$, EDM aim to learn the parameter $\theta$ by minimizing the KL divergence between $\rho_{D}$ and $\rho_{\pi^\theta}$, $D_{KL}(\rho_D||\rho_{\pi^\theta})$, where $\rho_{D}$ and $\rho_{\pi^\theta}$ represent the occupancy measures for the expert trajectories and for the induced policy $\pi^\theta$, respectively.
Despite its success in \emph{offline} learning, EDM assumes that all demonstrations adhere to a \emph{single} reward function, which is often impractical for human-centric tasks in the real world, such as education, where students may have multiple reward functions evolving over time. To address the challenge of \emph{multiple} reward functions \emph{varying across} demonstrations, an Expectation-Maximization(EM)-EDM \cite{islam2024generalized} was introduced, iteratively clustering demonstrations in the \textit{E-step} and inducing the policy for each cluster by EDM in the \textit{M-step}. The input to EM-EDM will be the sub-trajectories $\{\mathcal{D}^{\hat{j}} | \hat{j}=1,...,\hat{N}\}$, where $\hat{N}$ is the total number of sub-trajectories learned from \emph{Sub-trajectory partitioning}. Finally, THEMES yields sub-trajectory clusters with respective policies $\{\pi_{o} | o \in [1,O]\}$ as output, representing multiple student pedagogical strategies evolving over time.

\vspace{-0.1in}
\subsection{Baselines}
\label{subsec:baselines}
\subsubsection{Behavior Cloning (BC)}
Behavior cloning \cite{syed2012imitation,raza2012teaching} is a supervised learning approach used in AL where an agent learns to imitate an expert by directly mapping observations to actions based on the expert's demonstrations. The objective is to mimic the expert's behavior, hence, learn a policy $\pi_{\theta}(a|s)$ parameterized by $\theta$ that approximates the expert behavior: $L(\theta) = \frac{1}{n} \sum_{i=1}^{n} \lVert \pi_{\theta}(a_i|s_i) - a_i \rVert^2$.
It is important to note that while BC is straightforward, it can be sensitive to distribution shifts and compounding errors, especially when the learned policy deviates from the expert demonstrations.
\vspace{-0.2in}
\subsubsection{Gaussian Process (GP) + Deep Q-Network (DQN)}
To infer immediate rewards from the final delayed reward of each trajectory, this method \cite{azizsoltani2018adaptive} first employs GP to model a distribution function $f$ that captures the expected values and standard deviations of the unknown intermediate rewards. The inference is formulated as a minimum mean square error (MMSE) estimation problem, where the additive Gaussian rewards are estimated such that their sum approximates the observed delayed return \cite{guo2005mutual}. This approach assumes that rewards at each timestep follow a Gaussian distribution and that the total reward across the trajectory equals the observed delayed outcome. Once immediate rewards are inferred, a Deep Q-Network (DQN) is trained on the reconstructed immediate reward-labeled trajectories. 
% DQN approximates the optimal action-value function $Q^*(s, a)$ using the Bellman equation:
% \begin{equation}
% Q^*(s, a) = \mathbb{E}_{s'} \left[ r + \gamma \max_{a'} Q^*(s', a') \right],
% \end{equation}
% where $r$ is the inferred immediate reward, $\gamma$ is the discount factor, and $s'$ is the next state. The Q-network is trained by minimizing the squared temporal-difference (TD) error between both sides of this equation. 
This enables the agent to learn a policy that maximizes the expected cumulative reward over the reconstructed trajectories.
\vspace{-0.2in}
\subsubsection{Multi-modal Imitation Learning (MIL)}
MIL \cite{hausman2017multi} tackles imitation learning from unstructured and unlabeled demonstrations by learning a multi-modal policy. Instead of assuming demonstrations are pre-segmented by task, MIL jointly segments skills and learns their corresponding policies. The method augments the standard policy $\pi(a|s)$ with a latent intention variable $i \sim p(i)$, resulting in a conditional policy $\pi(a|s, i)$. The intention variable $i$ selects a specific mode of the policy corresponding to a distinct skill. MIL optimizes a generative adversarial imitation learning objective, extended with an additional latent intention loss, encouraging the policy to maximize the mutual information between the latent intention and the generated behaviors.
% \begin{align}
% \max_{\theta} \min_{w} \quad 
% & \mathbb{E}_{i \sim p(i), (s,a) \sim \pi_{\theta}} \left[ \log D_w(s,a) \right] 
% + \mathbb{E}_{(s,a) \sim \pi_E} \left[ \log (1 - D_w(s,a)) \right] \nonumber \\
% & + \lambda_I \mathbb{E}_{i, (s,a)} \left[ \log p(i|s,a) \right]
% + \lambda_H \mathcal{H}(\pi_{\theta}(a|s)),
% \end{align}
% where $D_w$ is the discriminator, $p(i|s,a)$ predicts the latent intention from state-action pairs, and $\mathcal{H}$ denotes entropy regularization. 
Through this framework, MIL simultaneously discovers the distinct skills present in demonstrations and learns to imitate them using a single, unified policy. 
% Although originally proposed for online settings, MIL can be adapted to offline scenarios by training entirely from a fixed demonstration dataset.
\vspace{-0.2in}
\subsubsection{Adapted Hierarchical Inverse Reinforcement Learning (AHIRL)}
HIRL \cite{krishnan2016hirl} addresses the challenge of long-horizon tasks with delayed rewards by segmenting demonstrations into a sequence of locally linear sub-tasks. Each sub-task is defined by consistent changes in local linearity, detected using a Bayesian nonparametric Gaussian Mixture Model (DP-GMM) over a featurized trajectory space.
% Given $G = [\rho_1, \ldots, \rho_k]$ denote the sequence of inferred sub-goal regions. 
To ensure sequential execution, HIRL augments the original state-space $\mathcal{S}$ with a binary vector $v \in \{0,1\}^Q$ that tracks progress across $Q$ sub-goals. 
% The augmented MDP becomes $\mathcal{M}_H = (\mathcal{S} \times \{0,1\}^k, \mathcal{A}, P', R, T)$. 
A reward function $R_\theta(s, a, v)$ is then learned using Maximum Entropy IRL over this augmented space. This enables the agent to reason about both current state and sub-task completion history. 
% In our offline adaptation, we reuse the segment clustering and local reward structure, but employ EDM-based policy learning rather than online RL, enabling HIRL to operate effectively in a strictly offline setting.

%%%%%%%%%%%%%%%%%%%%%%%%%%%%%%%%%%%%%%%%%%%%%%%%%%%%%%%%%%%%%%%%%%%%%%%%%%%%%%%%%%%%%%%%%%%%%%%%%%%%
\vspace{-0.1in}
\section{Experiment}
% We gathered our data by letting undergraduate students from the same major and same year complete an online intelligent tutoring system (ITS) (Figure \ref{fig:pyrenees-interface}) that introduced them to probability concepts like Bayes' Theorem and Addition Theorem. Using training problems, the students were guided through the instruction. The tutor gave detailed instructions, quick responses, and on-demand assistance for every problem. The assistance was given in the form of progressively more detailed hints. The bottom-out hint, the final one in the sequence, gave the students precise instructions. Students could decide throughout training how to solve the next stage pedagogically by either working through it alone, watching the tutor work through it, or doing it collaboratively. The tutor will ask questions to get the answer if they want to solve it independently; if not, the instructor will show or tell.
\vspace{-0.1in}
This study utilized a Probability ITS (Figure \ref{fig:pyrenees-interface}) deployed in the Spring and Fall semesters of a large undergraduate STEM course at a public university. Designed by domain experts and overseen by departmental committees, the system aims to teach entry-level undergraduates ten key probability principles, \textit{e.g.}, complement theorem and Bayes' rule, through solving a series of complex problems, each of which requires 30 to 60 interactions.

%the system has been utilized by over 3,000 students, generating approximately 1 million recorded interaction logs over 11 academic years.

% \begin{figure}
%     % \vspace{-0.2in}
%     \centering
%     \includegraphics[scale=0.3]{Figures/Pinterface.png}
%     \caption{Interface of the probability ITS}
%     % \caption{Interface of the probability ITS. The problem statement window (\textit{top}) presents the descriptions of the problem. The dialog window (\textit{middle right}) shows the message the tutor provides to the students. Responses, e.g., writing an equation, are entered in the response window (\textit{bottom right}). Any variables and equations generated by this process are shown on the variable window (\textit{middle left}) and equation window (\textit{bottom left}).}
%     \label{fig:pyrenees-interface}
%     \vspace{-0.2in}
% \end{figure}

\vspace{-0.15in}
\subsubsection{Data Collection}
Data were collected from 221 students over four semesters: 67 in Spring 2021 (S21), 56 in Spring 2022 (S22), 54 in Spring 2024 (S24), and 44 in Fall 2024 (F24). All students followed a standardized process, including the same textbook, pre-test, training, and post-test within our ITS. During the textbook, all students studied the 10 probability principle, read a general explanation, and worked through various single- and multi-principle problems. The pre-test and post-test comprised 8 and 12 open-ended complex problem-solving questions, respectively. % Students received no feedback on their responses and were not allowed to revisit previous questions during either phase. 

During ITS training, all students were presented with the same 12 training problems in a fixed sequence. Each problem required applying 3 to 11 domain principles for a solution. For two of these problems, all students were required to solve them independently to familiarize themselves with the ITS interface. For the remaining \textbf{10} problems, \textbf{students would decide among three approaches}: \emph{1) solving the problem independently}, \emph{2) collaborating with the ITS}, or \emph{3) viewing a worked-out example}.  Students' pre-test and post-test answers were graded independently by two evaluators using a double-blind process, with discrepancies resolved through discussion to reach a final grade. All test scores were normalized to the range of [0 - 100] for comparison purposes. Student data were obtained anonymously through an exempt IRB-approved protocol. 
\vspace{-0.1in}
\subsubsection{130 State Feature Representation}
Given that students' underlying learning status are unobservable \cite{mandel2014offline}, two domain experts define the \textit{\textbf{State}} space through \textbf{130}-dimensional continuous features including: \textbf{\textit{10} Autonomy} features that measure the student's work, such as the number of elicits since the last tell; \textbf{\textit{22} Temporal} features that represent time-related information, like average time per step; \textbf{\textit{31} Problem Solving} features that reflect contextual information about problem-solving, including difficulty level; \textbf{\textit{57} Performance} features that indicate student's performance so far, such as the percentage of correct responses; \textbf{\textit{10} Hints} features that show data on hint usage, such as total hints requested.

\vspace{-0.1in}
\subsubsection{Expert Student Demonstrations}
\label{TrajSelection}
\begin{figure}[htbp]
    \centering
    \begin{minipage}[b]{0.55\textwidth}
        \centering
        \includegraphics[scale=0.2]{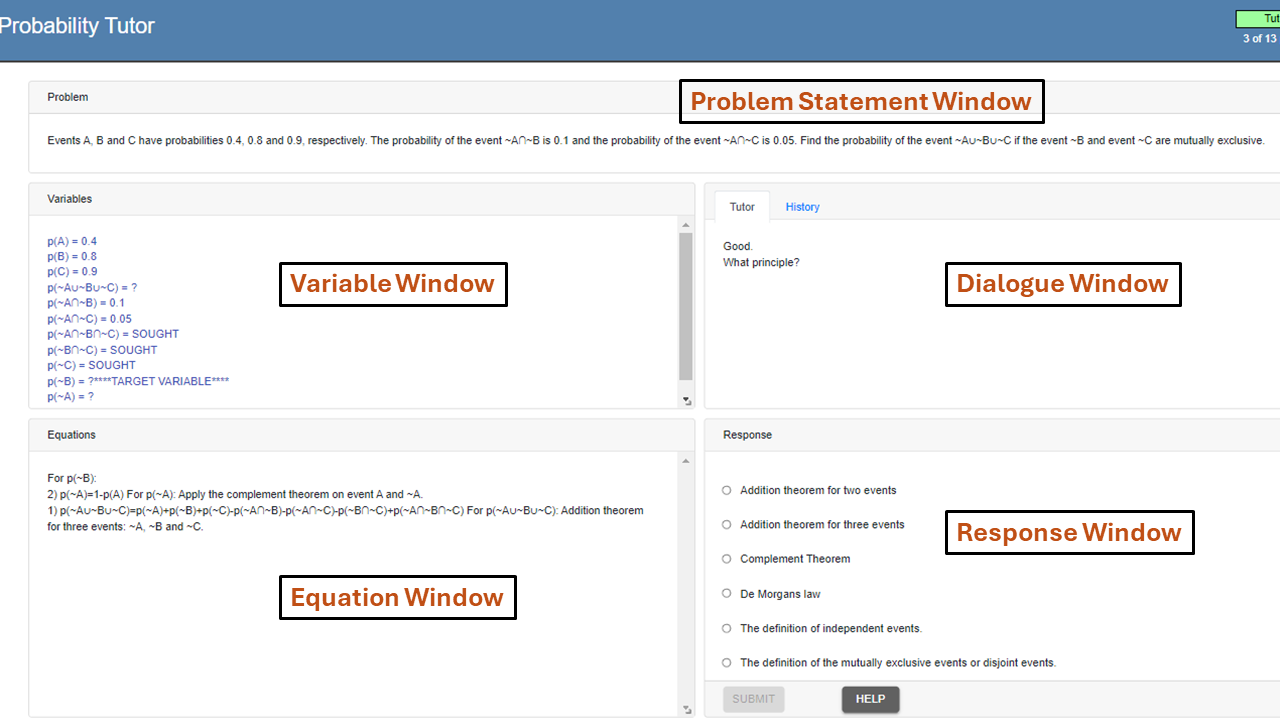}
        \caption{ITS Interface}
        \label{fig:pyrenees-interface}
    \end{minipage}
    \hfill
    \begin{minipage}[b]{0.42\textwidth}
        \centering
        \includegraphics[scale=0.2]{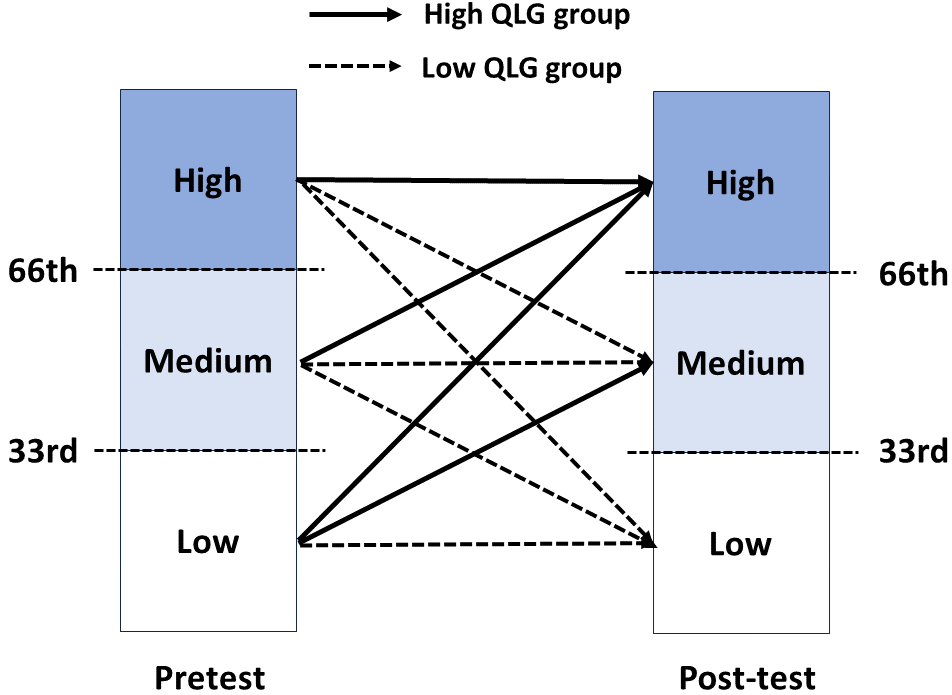}
        \caption{Quantized Learning Gain}
        \label{fig:qlg}
    \end{minipage}
    \vspace{-0.2in}
\end{figure}

To select demonstrations with optimal or near-optimal student interactions with the ITS, we used the Quantized Learning Gain (QLG) measure \cite{mao2018tracing}. Students were categorized into low, medium, or high-performance groups based on their pre- and post-test scores (as shown in Figure \ref{fig:qlg}). A "High" QLG is assigned to students who improved or remained in the high-performance group, while a "Low" QLG is given to those who dropped to a lower group or stayed in low/medium groups. This helped identify students who benefited most from our ITS. A total of 89 student trajectories from 4 semesters were used as expert demonstrations: 18 from S21, 24 from S22, 23 from S24, and 24 from F24.
% (Table \ref{tab:qlg_stat}). 

% To enhance the selection of high-quality demonstrations to induce more accurate yet effective AL policies, it is typically believed that the experts are performing the demonstrations in an optimal or near-optimal manner \cite{abbeel2004apprenticeship}. 
% Our original dataset contains 128 students from two semesters, Spring 21 (67 students) and Spring 22 (61 students). Each student spent \~ 2 hours on the probability ITS and completed around 400 steps.
% To select higher-quality trajectories from all students' interaction with the ITS, we use a qualitative measurement called Quantized Learning Gain (QLG) \cite{lin2017comparisons}, which is a binary qualitative measurement of students' learning gains from the pretest to the posttest to determine whether a student has benefited from a learning environment \cite{mao2018tracing}.

% \begin{figure}
%     \centering
%     \includegraphics[scale=0.2]{Figures/QLG.png}
%     \caption{Quantized Learning Gain}
%     \label{fig:qlg}
%     \vspace{-0.2in}
% \end{figure}

\vspace{-0.1in}
\subsubsection{Experimental Settings}
We select six competitive AL baselines for comparison against THEMES, as discussed in Section \ref{subsec:themes} and \ref{subsec:baselines}. The settings for the baselines are summarized as follows:\vspace{0.08in}

\noindent \textbf{1) BC} \cite{gleave2022imitation,syed2012imitation}: We use the BC implementation available in the \emph{imitation} \cite{gleave2022imitation} library and extend it to a customized environment corresponding to our ITS setting.

\noindent \textbf{2) GP+DQN} \cite{azizsoltani2018adaptive}: For this method, we employ Gaussian Process (GP) to infer immediate rewards first from the delayed final outcomes, e.g., learning gain of students for ITS setting, followed by a standard Deep Q-Network (DQN) implementation \cite{fan2020theoretical} for policy learning.

\noindent \textbf{3) EDM} \cite{jarrett2020strictly}: Since EDM is strictly offline, we minimally adapt the original implementation to fit the ITS setting.
% EDM represents the state-of-the-art in offline AL. It has been shown to outperform many recent AL methods using a \emph{single} reward function, and therefore we omit redundant comparisons with those baselines.

\noindent \textbf{4) Multi-modal Imitation Learning (MIL)} \cite{hausman2017multi}: MIL leverages InfoGAN to jointly learn sub-trajectories and policies. Although originally designed for online settings, we adapt it for offline learning.

\noindent \textbf{5) Adapted Hierarchical Inverse Reinforcement Learning (AHIRL)} \cite{krishnan2016hirl}: In our offline adaptation of HIRL, we reuse the segment clustering and local reward structure but employ EDM-based policy learning rather than online RL, enabling HIRL to operate effectively in a strictly offline setting.
% AHIRL clusters sub-trajectories using a Gaussian Mixture Model (GMM) and learns cluster-wise policies via maximum entropy IRL. To support offline training, we modify it by integrating EDM for policy induction.

\noindent \textbf{6) EM-EDM} \cite{islam2024generalized}: We extend the implementation of EDM to EM-EDM by incorporating the EM algorithm and maximizing the log-likelihood of cluster assignments as described in \cite{islam2024generalized}.

In addition to the six AL baselines, we evaluate the performance of THEMES against its two ablations. The first one is: \textbf{THEMES$_0$}, where the high-level states are obtained through sub-trajectory partitioning \cite{yang2021multi}, but the high-level actions are induced by EDM \cite{jarrett2020strictly}. The second one is: \textbf{THEMES$_1$}, where the high-level states are also learned by sub-trajectory partitioning, but high-level actions are induced by EM-EDM \cite{babes2011apprenticeship}. Finally, in \textbf{THEMES}, we utilize reward regulated sub-trajectory clustering to partition trajectories with hierarchically learned reward regulator and induce policies over the partitioned sub-trajectories via EM-EDM. For THEMES, the hyperparameters in sub-trajectory partitioning are determined based on the Bayesian Information Criteria (BIC) \cite{friedman2001elements}, with the number of sub-trajectory clusters $Q$ set to 6, window size $\omega$ set to 2, and the coefficients for sparsity and consistency terms set to 1e-3 and 4, respectively. For EM-EDM, the optimal number of clusters $O$ was determined heuristically as 3, by iteratively implementing the EM algorithm until either empty clusters are generated or the log-likelihood of the clustering results varies by less than a predefined threshold. The iterations for the overall THEMES framework are set as 10, since the clustering likelihood for both sub-trajectory clustering and EM-EDM converge within 10 iterations. 

\vspace{-0.2in}
\subsubsection{Experiment Setup and Evaluation Metrics}
It is important to emphasize that all models are evaluated using semester-based temporal cross-validation (3-fold) in this task, which only applied data from previous semesters individually or combined for training and is a much stricter approach than the standard cross-validation \cite{mao2020time}. We report Accuracy, Recall, Precision, F1 Score, AUC (Area Under the ROC Curve), APR (Area Under the Precision-Recall Curve), and Jaccard score. The Jaccard score measures the similarity between the predicted and true sets of labels, calculated as the size of the intersection divided by the size of the union of the two sets. Given the nature of our task, we consider AUC and Jaccard score to be the most critical metrics, as they are generally robust.
% since accuracy is the measure that is most widely acknowledged and AUC is thought to be generally more robust.
\vspace{-0.1in}
\section{Results}
\label{sec:results}
\vspace{-0.1in}

\renewcommand{\arraystretch}{1.3} 
\setlength{\tabcolsep}{0.5 pt} %%% Column size can be adjusted HERE %%% 
\begin{table*}
\vspace{-0.3in}
\caption{\centering Comparing THEMES with baselines and ablations}
\label{tab:result_xsem}
% \footnotesize
\small
    \centering
    \begin{tabular}{c|ccccccc} \hline
        Methods & Acc & Rec & Prec & F1 & AUC & AP & JAC \\ \hline
        1. BC \tiny{\cite{gleave2022imitation}} & $.366_{\pm .03}$ & $.366_{\pm .03}$ & $.384_{\pm .04}$ & $.363_{\pm .03}$ & $.523_{\pm .03}$ & $.362_{\pm .01}$ & $.224_{\pm .02}$ \\ % \hline
        2.GP+DQN \tiny{\cite{azizsoltani2018adaptive}} & $.389_{\pm .06}$ & $.389_{\pm .06}$ & $.399_{\pm .10}$ & $.297_{\pm .05}$ & $.525_{\pm .06}$ & $.386_{\pm .03}$ & $.193_{\pm .04}$ \\
        3. EDM \tiny{\cite{jarrett2020strictly}}& $.656_{\pm .05}$ & $.656_{\pm .05}$ & $.686_{\pm .03}$ & $.658_{\pm .05}$ & $.801_{\pm .03}$ & $.662_{\pm .04}$ & $.506_{\pm .05}$\\  %\hline
        4. MIL \tiny{\cite{hausman2017multi} } & $.508_{\pm .06}$ & $.508_{\pm .06}$ & $.486_{\pm .11}$ & $.456_{\pm .06}$ & $.627_{\pm .06}$ & $.465_{\pm .06}$ & $.319_{\pm .06}$\\
        5. AHIRL \tiny{\cite{krishnan2016hirl}}& $.545_{\pm .07}$ & $.545_{\pm .07}$ & $.566_{\pm .07}$ & $.545_{\pm .07}$ & $.688_{\pm .07}$ & $.537_{\pm .07}$ & $.388_{\pm .07}$\\ 
        6. EM-EDM \tiny{\cite{babes2011apprenticeship}} & $\mathbf{.705_{\pm .06}}$ & $\mathbf{.705_{\pm .06}}$ & $\mathbf{.725_{\pm .04}}$ & $\mathbf{.709_{\pm .05}}$ & $\mathbf{.847_{\pm .03}}$ & $\mathbf{.724_{\pm .05}}$ & $\mathbf{.561_{\pm .07}}$\\ \hline
        7. THEMES$_0$ & $.638_{\pm .05}$ & $.638_{\pm .05}$ & $.671_{\pm .06}$ & $.642_{\pm .06}$ & $.783_{\pm .05}$ & $.643_{\pm .07}$ & $.487_{\pm .06}$ \\ 
        8. THEMES$_1$ & $.745_{\pm .04}$ & $.745_{\pm .04}$ & $.761_{\pm .03}$ & $.747_{\pm .03}$ & $.875_{\pm .02}$ & $.768_{\pm .03}$ & $.605_{\pm .04}$\\
        9. THEMES & $\mathbf{.763_{\pm .02}}^{*}$ & $\mathbf{.763_{\pm .02}}^{*}$ & $\mathbf{.773_{\pm .02}}^{*}$ & $\mathbf{.765_{\pm .02}}^{*}$ & $\mathbf{.879_{\pm .02}}^{*}$ & $\mathbf{.773_{\pm .04}}^{*}$ & $\mathbf{.626_{\pm .03}}^{*}$\\ \hline 
    \end{tabular}
    \vspace{-0.2in}
\end{table*}

\begin{figure}
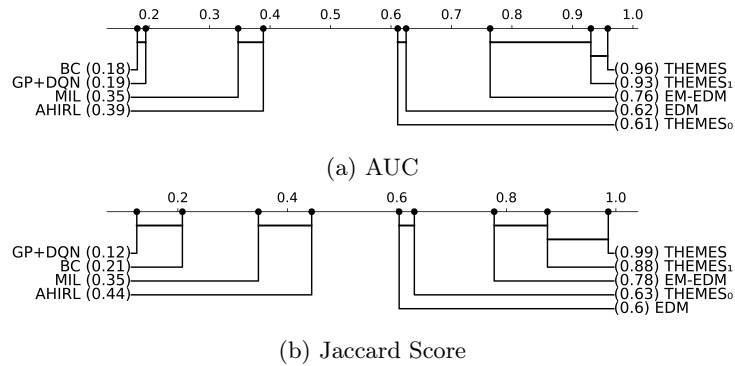

\vspace{-0.13in}
  \centering
  \begin{subfigure}{0.8\textwidth}
    \centering
    \includegraphics[width=\linewidth]{Figures/cd_auc.png}
    \caption{AUC}
    \label{subfig:cd_auc}
  \end{subfigure}
  \hfill
  \begin{subfigure}{0.8\textwidth}
    \centering
    \includegraphics[width=\linewidth]{Figures/cd_jaccard.png}
    \caption{Jaccard Score}
    \label{subfig:cd_jac}
  \end{subfigure}
  \caption{Critical difference diagram for AUC and Jaccard Score}
  \label{fig:cd_plot}
  \vspace{-0.2in}
\end{figure}

Table \ref{tab:result_xsem} reports the average and standard deviation of the overall semester-based temporal cross-validation results for the different models, with the best baselines and ablations highlighted in bold, and the overall best results are highlighted with *. The critical difference diagram, based on the Conover-Friedman tests ($p = 0.05$), for AUC and Jaccard, is presented in Figure \ref{fig:cd_plot}. The results clearly show that THEMES \emph{significantly} outperform all other methods in all evaluation metrics. Specifically, THEMES outperforms THEMES$_1$, demonstrating the effectiveness of hierarchically learned \emph{reward regulator} in incorporating decision-making patterns during sub-trajectory partitioning. Furthermore, the three THEMES-based methods consistently and significantly outperform competitive baselines such as AHIRL and MIL, which share the assumption of \emph{multiple} reward functions \emph{evolving} over time. This suggests that the effectiveness of high-level state representation is enhanced by incorporating \emph{time-awareness} during sub-trajectory partitioning. When comparing THEMES and THEMES$_{0}$ with EM-EDM, significant improvements are observed, highlighting the power of high-level state representation to capture time-varying patterns across partitioned sub-trajectories. Although EM-EDM is better than EDM, which supports the assumption of \emph{multiple} reward functions, the superior performance of THEMES over EM-EDM suggests that the assumption was insufficient to model time-varying decision-making. Moreover, THEMES demonstrates exceptional consistency between semesters, as evidenced by the minimal standard deviation in Table \ref{tab:result_xsem}. Notably, among all individual fold results of semester-based cross validation, we found that THEMES can achieve an AUC of 0.899 and a Jaccard of 0.653, using only \emph{18} trajectories of S21 to predict student strategies in S22. Finally, all of these results underscore the potential of THEMES as a sample-efficient alternative to capture evolving student pedagogical strategies.

\vspace{-0.2in}
\subsubsection{A Case Study}
\begin{figure}
\vspace{-0.2in}
    \centering
    \includegraphics[scale=0.43]{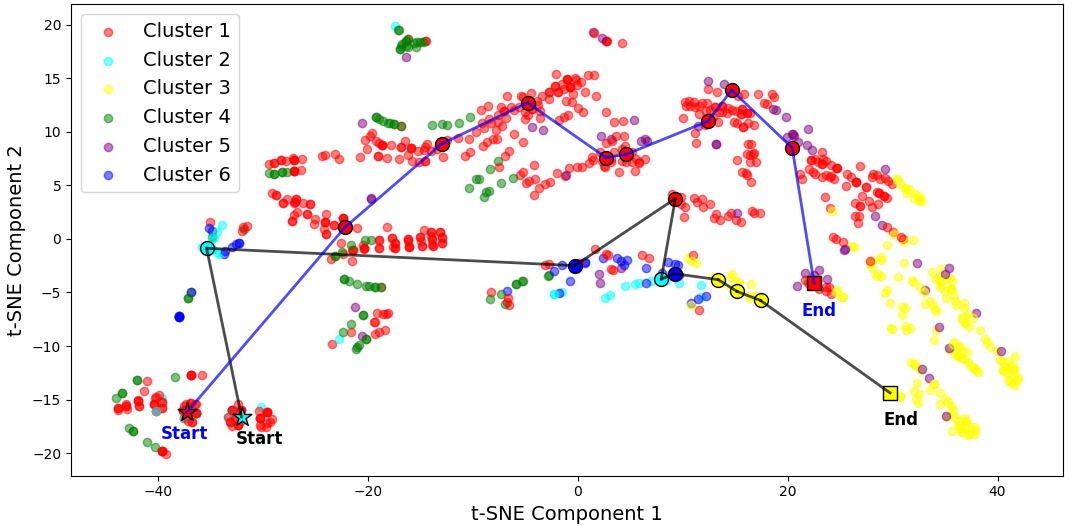}
    \caption{t-SNE Visualization of Expert Student Demonstrations—\fcolorbox{lightblue}{lightblue}{blue} Trajectory (representing a fixed reward function) vs. \fcolorbox{lightgray}{lightgray}{black} Trajectory (indicating evolving reward functions)}
    \label{fig:casestudy}
    \vspace{-0.3in}
\end{figure}

Figure \ref{fig:casestudy} presents a t-SNE visualization of the original 130-state space from all expert student demonstrations. To facilitate better visualization, we applied t-distributed stochastic neighbor embedding (t-SNE) to project the data into a lower-dimensional space \cite{hinton2002tsne}. In this figure, different colors correspond to the six clusters identified by sub-trajectory clusters or high-level states in THEMES framework. The \fcolorbox{lightblue}{lightblue}{blue} and \fcolorbox{lightgray}{lightgray}{black} lines represent the expert student demonstrations from two distinct groups. The \fcolorbox{lightblue}{lightblue}{blue} trajectory remains consistently within sub-trajectory cluster 1, indicating a student likely driven by a single, stable intention throughout their learning process. In contrast, the \fcolorbox{lightgray}{lightgray}{black} trajectory begins in cluster 2 but evolves over time, transitioning through clusters 6, 1, 2, 6, and 3. This suggests that the latter student’s learning is guided by multiple, shifting intentions. AL methods like EDM, which assume a single reward function for each demonstration, can effectively capture the former trajectory but may struggle to account for the dynamic nature of the latter. In contrast, THEMES leverages sub-trajectory partitioning and clustering, enabling it to capture and adapt to these evolving intentions, resulting in a more accurate policy for the latter trajectory.

%Since underlying learning stages of students are unobservable \cite{mandel2014offline}, it is very challenging to visualize how incorporating time-awareness improves modeling student decision-making. However, we present a case study and show students' behavior with evolving intentions over time throughout the state space (Figure \ref{fig:casestudy}). Given that the state space dimension is as high as 130 in our case, we applied t-distributed stochastic neighbor embedding (t-SNE) to project the data into a lower-dimensional space for better visualization \cite{hinton2002tsne}. We plot all the expert student trajectories from a particular semester grouped into 6 sub-trajectory clusters derived from RMT-TICC in THEMES.

% As the sub-trajectory clusters were originally formed in a 130-dimensional space, capturing high-dimensional structures and relationships, when visualized using t-SNE in a 2D space, points belonging to the same cluster may not remain spatially cohesive. This is expected, as t-SNE is a non-linear dimensionality reduction technique that preserves local structure but does not always maintain global separability, especially when projecting highly complex, high-dimensional manifolds into two dimensions. Our case study involves two randomly sampled expert student trajectories.

%The \fcolorbox{lightblue}{lightblue}{blue} trajectory remains in the sub-trajectory cluster 1 throughout the whole time. In contrast, the \fcolorbox{green}{green}{green} trajectory begins from cluster 2, evolves over time and moves to cluster 6, 1, 6, and 3 sequentially. This suggests that the former student is likely to be driven by a single intention, while the latter student is driven by multiple evolving intentions over time. AL Methods such as EDM, which assumes that each demonstration is driven by a single reward function, can handle the former trajectory well but may not capture the evolving nature exhibited by the latter trajectory. On the other hand, THEMES can utilize the sub-trajectory partitioning and clustering to extract the high-level states in the later trajectory to yield a more accurate policy. 

%%%%%%%%%%%%%%%%%%%%%%%%%%%%%%%%%%%%%%%%%%%%%%%%%%%%%%%%%%%%%%%%%%%%%%%%%%%%%%%%%%%%%%%%%%%%%%%%%%%%
% \vspace{-0.1in}
\section{Caveats, Discussion \& Conclusion}
\vspace{-0.13in}
We leverage a generalized time-aware hierarchical AL framework, THEMES, to induce pedagogical policies from expert student demonstrations, with the assumption that multiple reward functions evolve over time. By integrating time-awareness with hierarchically learned reward patterns, THEMES enables more precise sub-trajectory partitions, resulting in highly effective pedagogical policies that align closely with expert student decisions. However, a limitation of this work lies in the relatively limited interpretation of evolving multiple reward functions in student demonstrations. Furthermore, this study is focused on a single ITS, although the time-aware hierarchical AL approach holds significant promise for other e-learning environments. Using student-ITS interaction logs, it can potentially validate the presence of multiple evolving intentions in diverse contexts. Moreover, while the trajectory lengths in our ITS were shorter than those in many real-world human-centric environments, THEMES still outperformed other approaches. Future work exploring e-learning environments with longer trajectories may reveal even more impressive results, further solidifying the framework's effectiveness in complex, evolving pedagogical settings.

%In this paper, we introduce a generalized time-aware hierarchical AL framework to induce student pedagogical strategies from expert demonstrations where \emph{multiple} reward functions \emph{evolve} over time. By incorporating both time-awareness and decision-making patterns encoded in hierarchically learned rewards, THEMES enables more fine-grained sub-trajectory partitions. These sub-trajectories further lead to more accurate policies in ITSs. The experimental results show that THEMES can effectively model complex pedagogical decision-making processes, demonstrating its strength in capturing evolving student pedagogical strategies. One caveat of this work is the limited interpretation of the evolving multiple reward functions in the student demonstrations. Moreover, this study focuses on one ITS. There are more STEM e-learning environments where a time-aware hierarchical AL approach can leverage student-ITS interaction logs to validate multiple evolving intentions.
%In addition, the trajectory length in our ITS was comparatively shorter than in many other human-centric environments in the real world. Despite that, THEMES showed its strength through superior performance. Employing THEMES in environments with longer trajectories can show even better results. 
% Although our current analysis primarily focuses on discrete actions, future work will explore more complex continuous action spaces.

\section{Acknowledgments}
This material is based on work supported by the National Science Foundation AI Institute for Engaged Learning (EngageAI Institute) under Grant No. DRL-2112635, Generalizing Data-Driven Technologies to Improve Individualized STEM Instruction by Intelligent Tutors (2013502), Integrated Data-driven Technologies for Individualized Instruction in STEM Learning Environments (1726550),  CAREER: Improving Adaptive Decision Making in Interactive Learning Environments (1651909). Any opinions, findings, and conclusions expressed in this material are those of the authors and do not necessarily reflect the views of the NSF.
%%%%%%%%%%%%%%%%%%%%%%%%%%%%%%%%%%%%%%%%%%%%%%%%%%%%%%%%%%%%%%%%%%%%%%%%%%%%%%%%%%%%%%%%%%%%%%%%%%%%

\bibliographystyle{splncs04}
\bibliography{reference2}

\end{document}